\documentclass[letterpaper, 10 pt, conference]{ieeeconf}
\IEEEoverridecommandlockouts             % Needed to use the \thanks command
\usepackage{cite}
\usepackage{graphics} % for pdf, bitmapped graphics files
\usepackage{epsfig} % for postscript graphics files
\usepackage{times}
\usepackage{amsmath}
\usepackage{amssymb}
\usepackage{mathtools}
\usepackage[inkscapeformat=png]{svg}
\usepackage{algpseudocode}
\usepackage{algorithmicx}
\usepackage[linesnumbered, ruled, vlined]{algorithm2e}
\algnewcommand{\AND}{\algorithmicand}
\usepackage{booktabs}
\usepackage{multirow}
\usepackage{tabularx}
\usepackage{url}
\usepackage{subfigure}
\usepackage{balance}

\title{\LARGE \bf Subassembly to Full Assembly: Effective Assembly Sequence Planning through Graph-based Reinforcement Learning}
\author{Chang Shu, Anton Kim, and Shinkyu Park% <-this % stops a space
\thanks{The work was supported by funding from King Abdullah University of Science and Technology (KAUST), and the SDAIA-KAUST Center of Excellence in Data Science and Artificial Intelligence (SDAIA-KAUST AI).}% <-this % stops a space
\thanks{The authors are with Electrical and Computer Engineering, King Abdullah University of Science and Technology (KAUST), Thuwal, 23955-6900, Kingdom of Saudi Arabia. Emails:
        {\tt\small \{chang.shu, anton.kim, shinkyu.park\}@kaust.edu.sa }}%
}

\begin{document}
\maketitle
\thispagestyle{empty}
\pagestyle{empty}
%%%%%%%%%%%%%%%%%%%%%%%%%%%%%%%%%%%%%%%%%%%%%%%%%%%%%%%%%%%%%%%%%%%%%%%%%%%%%%%%
\begin{abstract}
This paper proposes an assembly sequence planning framework, named Subassembly to Assembly (S2A). The framework is designed to enable a robotic manipulator to assemble multiple parts in a prespecified structure by leveraging object manipulation actions. The primary technical challenge lies in the exponentially increasing complexity of identifying a feasible assembly sequence as the number of parts grows. To address this, we introduce a graph-based reinforcement learning approach, where a graph attention network is trained using a delayed reward assignment strategy. In this strategy, rewards are assigned only when an assembly action contributes to the successful completion of the assembly task. We validate the framework's performance through physics-based simulations, comparing it against various baselines to emphasize the significance of the proposed reward assignment approach. Additionally, we demonstrate the feasibility of deploying our framework in a real-world robotic assembly scenario.

\end{abstract}

%%%%%%%%%%%%%%%%%%%%%%%%%%%%%%%%%%%%%%%%%%%%%%%%%%%%%%%%%%%%%%%%%%%%%%%%%%%%%%%%

\section{Introduction} \label{sec:introduction}
\begin{figure}
    \centering
    \includegraphics[width=0.9\linewidth, height=13em]{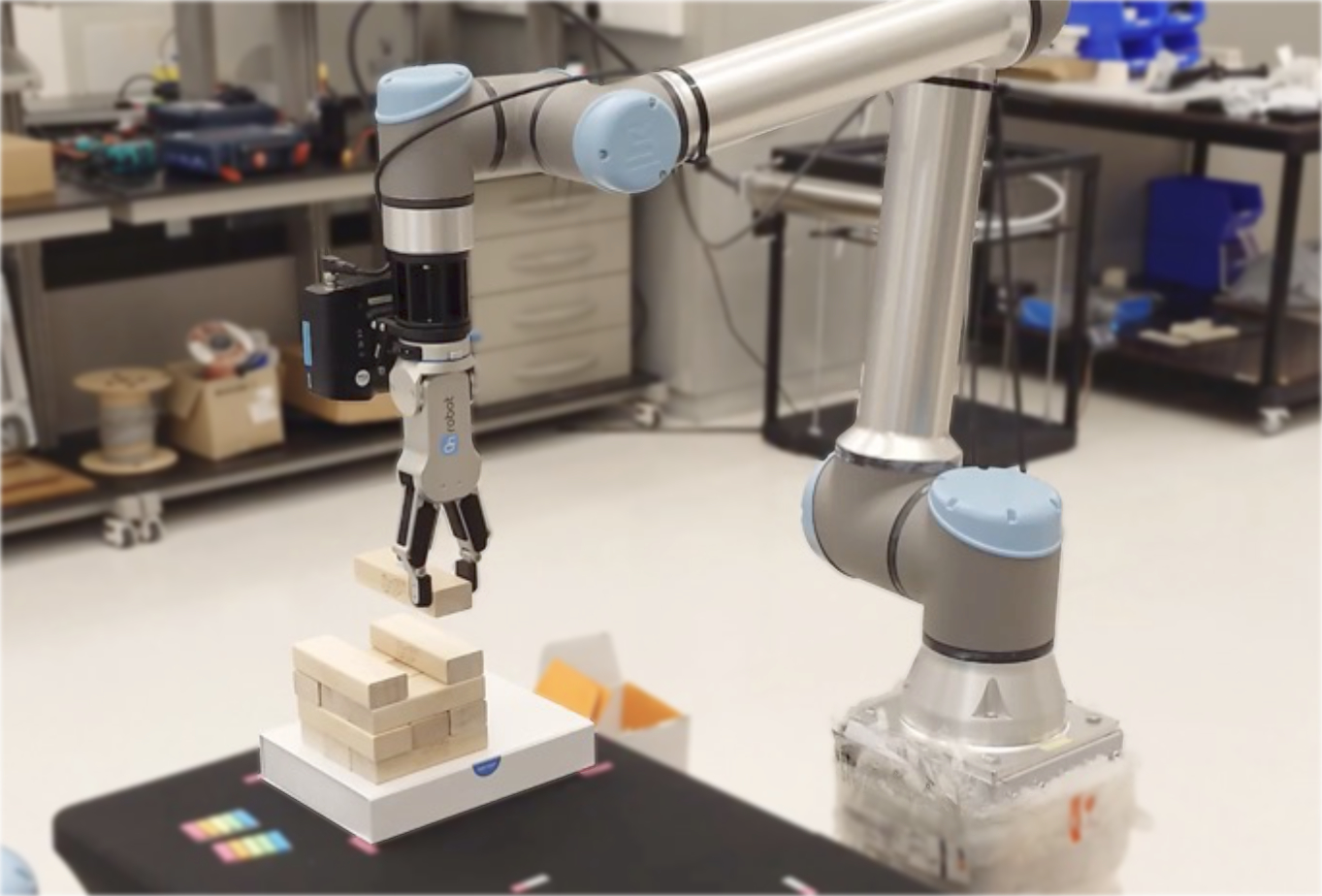}
    \caption{Demonstration of Jenga assembly using a robotic manipulator.}
    \label{fig:jenga_assembly}
    \vspace{-1em}
\end{figure}

An essential step towards assembly automation is assembly sequence planning (ASP), which aims to determine a feasible sequence of assembly operations that enables a manipulator to bring individual parts together. An effective assembly sequence planning framework infers precedence relationships between parts based on their geometric shapes, thereby avoiding unnecessary attempts at infeasible assembly operations. As illustrated in Fig.~\ref{fig:jenga_assembly}, the correct sequence for assembling a Jenga tower begins at the bottom and progress upward, taking into account the manipulator's kinematics and the influence of gravity.

Planning feasible assembly sequences for large-scale assemblies is a challenging task. One major issue ASP faces is \textit{combinatorial explosion} \cite{jimenez2013survey}, where the complexity of the problem grows exponentially with the number of parts. To address this, previous works \cite{10342352, 10160920, 10611595} have applied deep learning algorithms to predict feasible assembly sequences step-by-step. However, these methods only evaluate the feasibility of the current step when selecting the next part for assembly, neglecting the impact of future steps. For instance, parts already assembled might obstruct the assembly of remaining components.
Additionally, the proportion of feasible solutions to the ASP problem is extremely small due to numerous precedence constraints imposed by the complex geometries of assembly problem instances. This creates a significant imbalance between positive and negative samples, making it difficult to train effective assembly sequence planners.

We propose the \textit{Subassembly to Assembly} (S2A) framework for planning feasible assembly sequences over a directed graph. Our framework employs a Graph Neural Network (GNN) to capture the precedence relationships between assembly operations. The GNN is trained using a Q-learning method with a delayed reward assignment strategy that discourages assembly actions resulting in incomplete assemblies.

Our contributions are as follows: 1) We propose a learning-based approach for planning feasible assembly sequences over a directed graph, using a GNN trained with the Double Deep Q-Learning \cite{van_Hasselt_Guez_Silver_2016} method. 2) We design a delayed reward assignment strategy that considers not only the physical feasibility of an action but also its long-term impact on the remaining assembly process. 3) We incorporate physics-based simulation to evaluate the physical and kinematic constraints of the ASP problem, ensuring successful assembly execution in both simulation and real-world experiments.

The remainder of the paper is organized as follows: In Section~\ref{sec:formulation}, we present the formulation of the ASP problem, followed by a literature review in Section~\ref{sec:related-work}. Section~\ref{sec:method} details the S2A framework, built on a GNN architecture, and explains how the Q-learning method is employed to train the GNN. Extensive simulations are used to validate the proposed framework in Section~\ref{sec:experiments}, along with a demonstration on a robotic platform. We conclude the paper with a summary and discuss future research directions in Section~\ref{sec:conclusion}.

\section{Problem Description} \label{sec:formulation}
\begin{figure}
    \centering
    \subfigure[\label{fig:blocks}]{\includegraphics[width=0.7\linewidth]{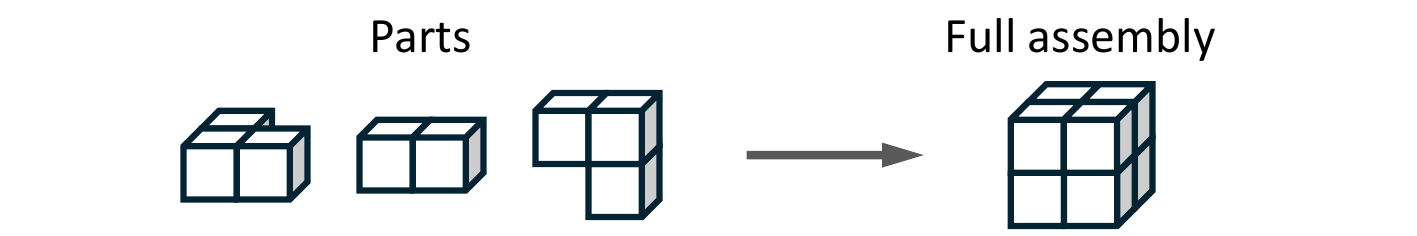}}
    \subfigure[\label{fig:graph}]{\includegraphics[width=0.7\linewidth]{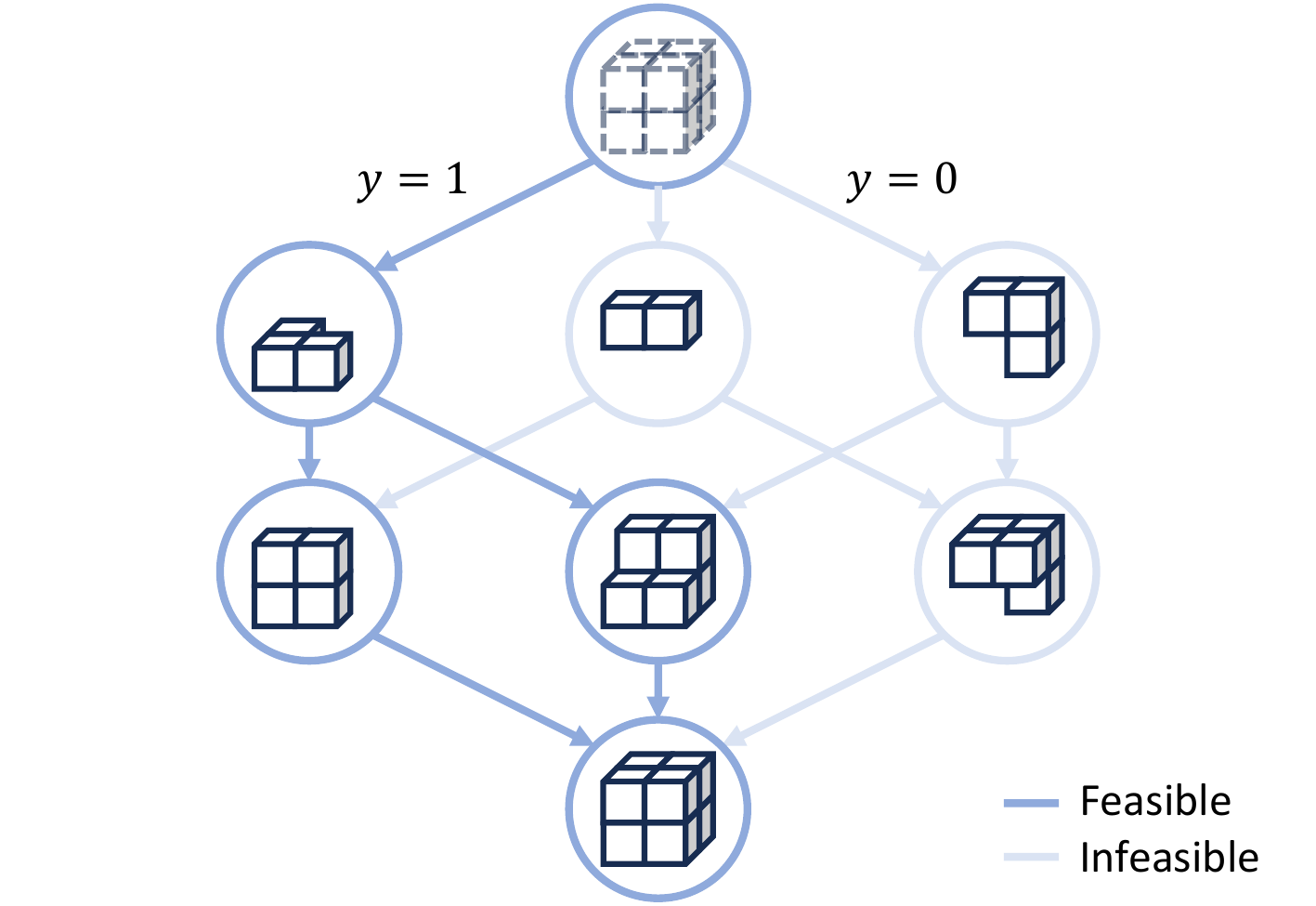}}
    \caption{(a) An example of assembling $3$ Tetris-like blocks, and (b) an illustration of an ASP problem represented as a directed graph.}
    \vspace{-1em}
\end{figure}

We focus on ASP for the block assembly task, where a set of $M$ Tetris-like blocks must be assembled into a pre-defined structure, as illustrated in Fig.~\ref{fig:blocks}. We refer to these building blocks as \textit{parts} and denote them as $\mathcal{P}=\{p_1, \dots, p_M\}$. Our objective is to determine a sequence of feasible assembly steps, as illustrated in Fig.~\ref{fig:graph}, that will enable a manipulator to execute collision-free motions to position all parts in their designated assembly locations. 

\subsection{Markov Decision Process Representation}
We represent the ASP problem using a Markov Decision Process (MDP), denoted as a tuple $(\mathcal{S}, \mathcal{A}, P, R, \gamma)$. Here, $\mathcal{S}$ is the state space, $\mathcal{A}$ is the action space, $P: \mathcal{S}\times \mathcal{A}\times \mathcal{S} \rightarrow [0, 1]$ is the transition probability, $R: \mathcal{S} \times \mathcal{A} \rightarrow \mathbb{R}$ is the reward function, and $\gamma \in [0, 1]$ is the discount factor. The state $s_t \in \mathcal{S}$ is defined to capture the current status of the assembly task, taking a value $v \in 2^{\mathcal{P}}$, where $2^{\mathcal{P}}$ denotes the power set of $\mathcal{P}$. We refer to each $v$, other than the full assembly state $\mathcal P$, as a \textit{subassembly}, representing a partially completed assembly of the parts. Additionally, we define the complement $\bar v$ of $v$ as $\bar v = \mathcal P \setminus v$, representing the set of remaining parts to be assembled. 

Given the current state $s_t = v$, the assembly action $a_t \in \mathcal{A}$ is defined as adding a new part $p \in \overline{v}$ to construct a larger subassembly $v \cup \{p\}$. Upon executing this action, the MDP generates a reward $r_t = R(s_t, a_t)$ and updates the current state to $s_{t+1} = v \cup \{p\}$ with a transition probability $P(s_{t+1} | s_t, a_t)$.
The MDP starts from the initial state $s_1 = \emptyset$ and repeatedly selects an assembly action until a termination criterion is met, such as $s_{t+1} = \mathcal{P}$. After $M$  consecutive actions are selected, the assembly sequence is represented as a list of visited assembly states $(s_1, \dots, s_{M})$. To guarantee the successful assembly of all parts, each action must comply with specific precedence constraints, ensuring the physical feasibility of the assembly process when using a manipulator.

\subsection{Feasibility of Assembly Actions}
Considering the manipulator's kinematics and the influence of gravity, we assert an assembly action $a_t$ is feasible if it satisfies the following two conditions: 1) there exists a collision-free motion trajectory that allows the manipulator to assemble a new part $p$ onto the current subassembly $s_t$ using the object manipulation actions -- grasping and pushing -- described in Appendix~A,
% \cite[Appendix~A]{chang_robotic_assembly}, 
and 2) the resulting state $s_{t+1}$ is \textit{structurally stable}, meaning the subassembly $s_{t+1}$ can sustain its structural integrity under the effects of gravity. 

In this work, we evaluate physical feasibility using physics-based simulation rather than relying solely on geometric reasoning. The kinematic constraint is assessed through sampling-based motion planning. Specifically, we use Rapidly-exploring Random Tree$^*$ (RRT$^*$) \cite{rrt} with a pre-defined search budget to compute an assembly trajectory that allows the manipulator to add $p$ to $s_{t}$. During the search, assuming the manipulator's long reach, we only integrate the gripper model for collision checking in the RRT$^*$ planning. If no such trajectory can be found within the search budget, the action $a_t$ is deemed infeasible. 

\begin{figure}
    \centering
    \includegraphics[width=.9\linewidth]
    {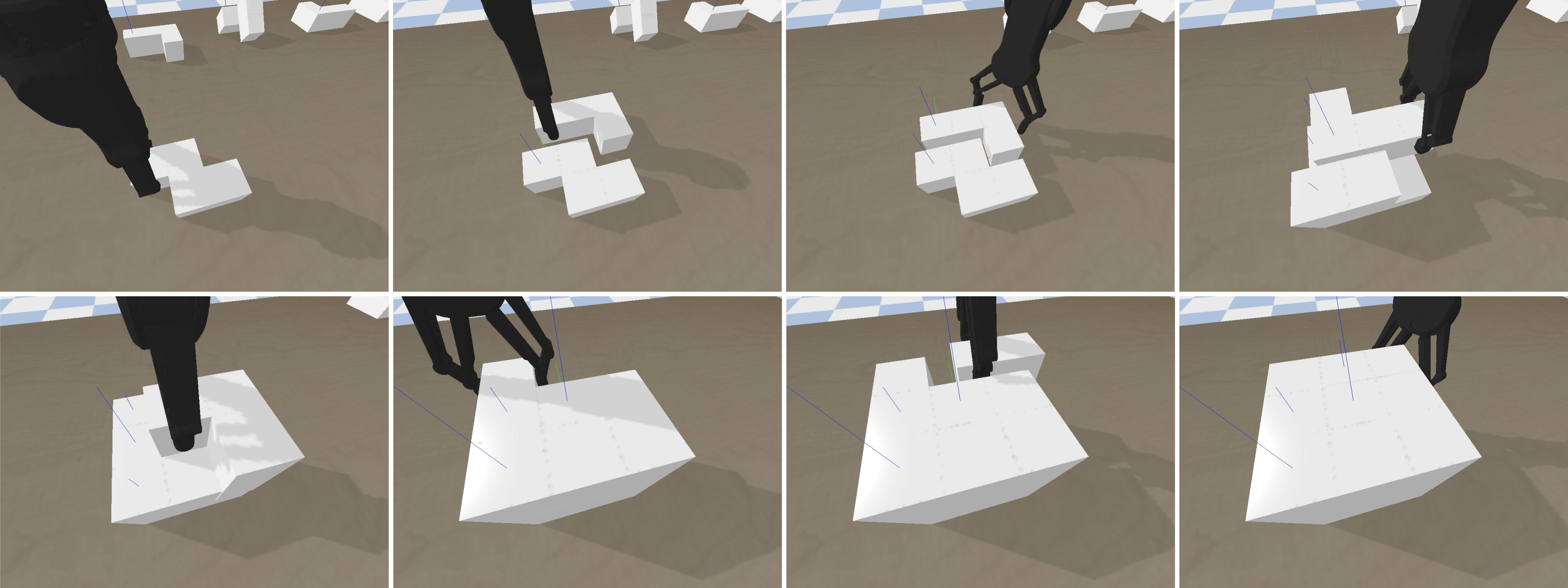}
    \caption{Pybullet simulation showcasing the assembly of a 7-part block.}
    \label{fig:pybullet-demo}
    \vspace{-1em}
\end{figure}

To assess the structural stability of the next subassembly $s_{t+1}$, we use PyBullet \cite{pybullet} to simulate whether the parts in $s_{t+1}$ can maintain their positions and orientations over an extended simulation period. If they do, the subassembly is considered stable.
Fig.~\ref{fig:pybullet-demo} illustrates the PyBullet environment, where the manipulator uses the RRT$^\ast$ planner to compute collision-free trajectories and carry out the assembly.

\subsection{Problem Formulation} 
We formulate the ASP as a path-finding problem on a weighted directed graph $\mathcal G = (\mathcal V, \mathcal E, \mathcal Y)$, as illustrated in Fig.~\ref{fig:graph}, where the nodes represent states and the edges correspond to actions within the MDP $(\mathcal{S}, \mathcal{A}, P, R, \gamma)$. In particular, the node set $\mathcal{V} = \{v_1, \dots, v_N\}$ represents the set of all $N=2^M$ subassemblies, with $v_1 = \emptyset$ as the root and $v_N = \mathcal{P}$ as the full assembly state. According to the assembly action definition, the out-neighbors of a node $v$, formally defined as $\mathcal{N}_{+} (v) =\{v \cup \{p\} \in \mathcal{V} \mid p \in \bar v\}$, represent potential next subassemblies after applying an assembly action at $v$ within the MDP. Thus, each assembly action $a \in \mathcal A$ is represented by an edge $(v, u)$, where ${u \in \mathcal{N}_{+}(v)}$. The set of all admissible assembly actions is denoted by $\mathcal{E} \subseteq \mathcal{V}\times \mathcal{V}$, with $|\mathcal{E}| = M \times 2^{M-1}$.\footnote{Admissible assembly actions, whether feasible or infeasible, involve adding a single part to an existing subassembly.} The set $\mathcal Y = \{y_{vu}\}_{(v,u) \in \mathcal E}$ contains the labels of all edges. To distinguish feasible actions from infeasible ones, each edge $(v, u) \in \mathcal{E}$ is assigned a label $y_{vu} \in \mathcal Y$, where $y_{vu} = 1$ indicates a feasible action and $y_{vu} = 0$ otherwise. 

Under this graph-based representation, the ASP is equivalent to finding a path over feasible edges from the root $v_1$ to the full assembly state $v_N$. During the assembly process, the MDP state $s_t$ must transition along a feasible edge to one of its out-neighbors, continuing until either reaching the full assembly $v_N$ or encountering a dead end, where the subassembly state has no feasible edges to its out-neighbors.

\section{Related Work} \label{sec:related-work}
ASP, commonly framed as a combinatorial optimization (CO) problem, is usually represented using graphs and solved with standard graph search algorithms \cite{jimenez2013survey}. One of the widely adopted representations is the AND/OR graph \cite{54734}, which models feasible subassemblies and their precedence relationships as a directed graph. However, traditional graph search algorithms become impractical for large-scale assembly problems due to the exponential growth in computational requirements. To mitigate this challenge, many efforts have been invested in designing evolutionary algorithms, such as ant colony optimization \cite{wang2005novel, wang2014mechanical}, genetic algorithms \cite{marian2006genetic}, and particle swarm optimization \cite{7797623}. 

More recently, deep learning approaches have been proposed to accelerate the ASP by recognizing underlying assembly rules, such as precedence relationships between parts \cite{9028142}, predicting the feasibility of each assembly step \cite{10342352, 10160920, 10611595}, or inferring assembly sequences \cite{10160424, 10611259} in an autoregressive fashion.
Unlike previous methods that rely on supervised learning paradigms, we propose addressing the ASP using model-free reinforcement learning (RL) applied to a directed graph. Our work is motivated by the recent success of RL in solving CO problems \cite{NIPS2017_d9896106, kool2018attention}. By learning Q-functions effectively, RL enables the robot to assess the feasibility of each assembly task, considering not only the current assembly state but also the potential to complete the entire assembly from that state. Additionally, graph searching techniques, such as beam search, allow the robot to explore multiple candidate solutions, increasing the likelihood of computing a feasible sequence plan at the cost of additional computation time.

RL methods have been employed to learn generalized assembly control policies in various simulated environments, such as furniture \cite{yu2021roboassembly} and magnetic blocks \cite{pmlr-v162-ghasemipour22a}. In contrast, our focus is on the sequence planning for robotic assembly, with validation through real-world experiments.
In \cite{9551620}, an RL method is proposed to guide the search for feasible assembly sequences, but it lacks generalizability.
The most closely related work is the block assembly presented in \cite{pmlr-v164-funk22a, 9981784}, where the goal is to approximate a 3D target geometry through the assembly of Tetris-like building blocks. In contrast, we tackle the ASP problem by generating a precise assembly sequence to achieve an exact final assembly. Furthermore, unlike \cite{pmlr-v164-funk22a, 9981784}, which use top-down grasping motions for assembly, we employ sampling-based motion planning to determine the robot gripper's trajectories, fully utilizing its 6D motion capabilities.

\section{Subassembly to Assembly (S2A) Framework} \label{sec:method}
\begin{figure*}
    \centering
    \includegraphics[width=\textwidth]{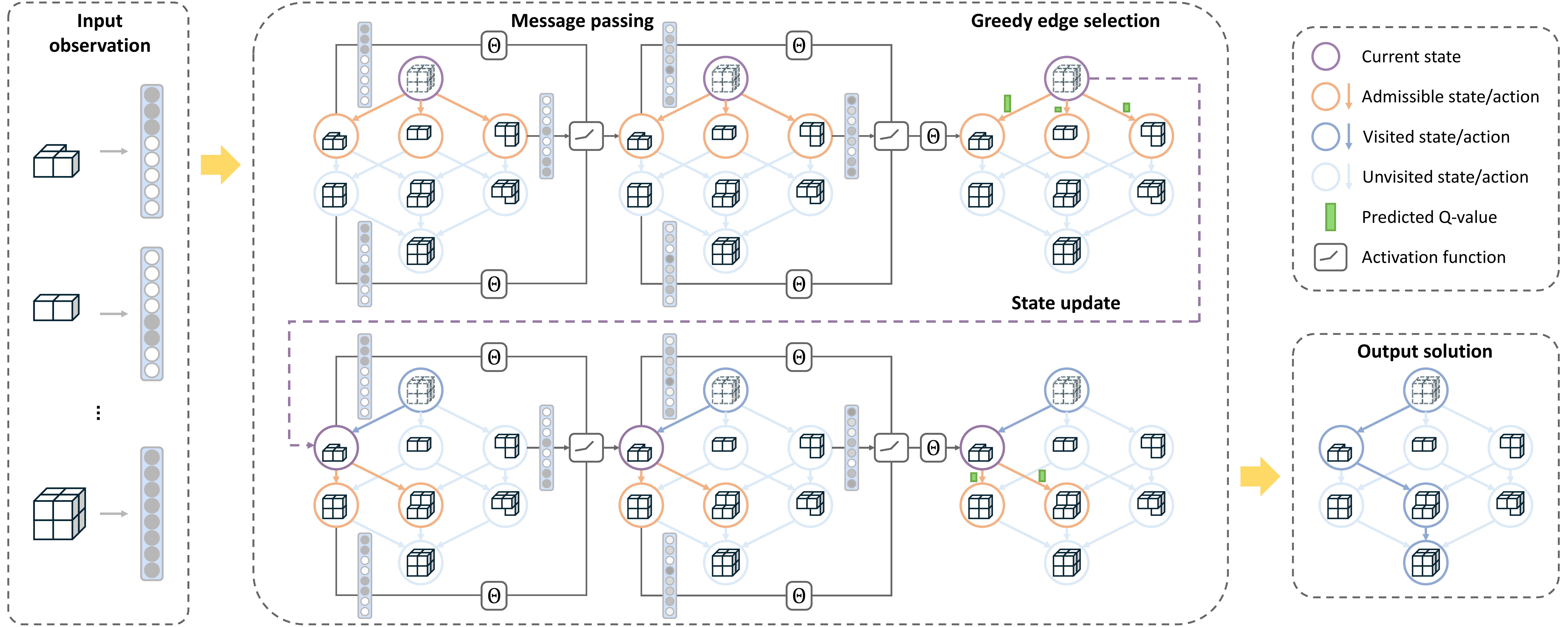}
    \caption{An overview of the S2A framework: Our framework takes $N$ volumetric representations of assembly states as input, applies $L$ rounds of message-passing procedures, and predicts Q-values to estimate the quality of admissible assembly actions. Starting from the root, a solution to the ASP problem is constructed by iteratively selecting the action with the highest Q-value.}
    \label{fig:overview}
    \vspace{-1em}
\end{figure*}
We propose the S2A framework, which employs Q-learning to estimate the quality of assembly actions across the directed graph $\mathcal{G}$.\footnote{As the name suggests, by learning the Q-function, our S2A framework enables the manipulator to complete assembly tasks starting from any subassembly state and progressing to the full assembly state. This design allows the manipulator to resume its task after interruptions or even restore a partially damaged assembly.} This learning process generates a Q-function that is subsequently used during the inference phase to guide the search for feasible assembly sequences. Below, we provide a comprehensive description of our S2A framework, detailing its components: assembly observations, GNN encoder, reward assignment strategy, and action selection policy. Fig.~\ref{fig:overview} illustrates our proposed framework.

\subsection{Graph Neural Network and Q-Function Design}
The S2A framework employs a GNN architecture to encode the ASP problem, where each node $v$ in the GNN corresponds to a subassembly and is associated with a vector encoding its geometric shape. To define this vector, we adopt a volumetric representation for all subassemblies in $2^{\mathcal P}$, similar to the approach in \cite{9981784}: The 3D assembly space is discretized into $d$ voxels, with the voxel occupancy captured by a $d$-dimensional feature vector $\mathbf{x}_v \in \mathbb{R}^{d}$. Each element $\mathbf{x}_v[i]$ is set to $1$ if the corresponding subassembly occupies the $i$-th voxel, and $0$ otherwise. The voxel grid resolution is selected to match the size of the unit cubes forming the individual parts.

Our GNN encoder is composed of $L$ layers of GATv2 \cite{brody2022how}, with each layer computing the embedding vector $\mathbf{h}_v^{(l)} \in \mathbb{R}^{f^{(l)}}$ of each node $v$ by attending over its out-neighbors. The embedding vector $\mathbf{h}_v^{(1)}$ of the first layer is initially set to $\mathbf{x}_v$. To determine the importance of each neighbor $u\in \mathcal{N}_{+}(v)$ and $v$ itself, GATv2 employs an attention mechanism to compute an attention coefficient $\alpha_{vu}^{(l)}$ for each edge $(v, u)$:
\begin{equation*}
    \alpha_{vu}^{(l)} \!=\! \frac{\exp{\left({\mathbf{a}^{(l)}} \cdot \sigma_1\!\left(\mathbf{\Theta}_1^{(l)} \mathbf{h}_v^{(l)} \!+\! \mathbf{\Theta}_2^{(l)} \mathbf{h}_u^{(l)}\right)\right)}}{\sum_{w \in \mathcal{N}_{+}(v) \cup \{v\}} \exp{\left({\mathbf{a}^{(l)}} \cdot \sigma_1\!\left(\mathbf{\Theta}_1^{(l)} \mathbf{h}_v^{(l)} \!+\! \mathbf{\Theta}_2^{(l)} \mathbf{h}_w^{(l)} \right)\right)}},
\end{equation*}
where $\mathbf{a}^{(l)} \in \mathbb{R}^{f^{(l)}}$ and $\mathbf{\Theta}_1^{(l)}, \mathbf{\Theta}_2^{(l)} \in \mathbb{R}^{f^{(l+1)} \times f^{(l)}}$ are learnable parameters and $\sigma_1$ is a nonlinear activation function ($\operatorname{LeakyReLU}$). Using these attention coefficients as weights, $\mathbf{h}_v^{(l)}$ is updated through the following message-passing procedure within its neighborhood:
\begin{equation*}
    \mathbf{h}^{(l+1)}_{v} = \sum_{u \in \mathcal{N}_{+}(v) \cup \{v\}} \alpha_{vu}^{(l)} \mathbf{\Theta}_2^{(l)} \mathbf{h}_u^{(l)}.
\end{equation*}

After $L$ iterations of the aforementioned message-passing procedure, we obtain an updated embedding vector $\mathbf{h}_v^{(L)}$ for each node $v$. These node embeddings are used to derive a parameterized Q-function, which estimates the likelihood that an assembly action, represented by an edge $(v, u)$ in $\mathcal G$, will result in the next feasible subassembly. Since valid assembly steps only exist between the current assembly state and its out-neighbors, we apply a mask to edges $(v,u)$ that do not satisfy $u \in \mathcal N_{+}(v)$. The Q-function is defined as follows:
\begin{equation*}
    Q(v,\! u) \!=\! \begin{cases}
    \mathbf{\Theta}_3^{(L)} \cdot \sigma_2 \left(\mathbf{\Theta}_4^{(L)} \left[\mathbf{h}^{(L)}_{v} \big\| \mathbf{h}^{(L)}_{u}\right]\right) & \text{if } u \!\in\! \mathcal{N}_{+} (v) \\
    -\infty & \text{otherwise,}
    \end{cases}
\end{equation*}
where $\mathbf{\Theta}_3^{(L)} \!\in\! \mathbb{R}^{f^{(L)}}$, $\mathbf{\Theta}_4^{(L)} \!\in\! \mathbb{R}^{f^{(L)} \times 2f^{(L)}}$ are learnable parameters, $\sigma_2$ is a nonlinear activation function ($\operatorname{ReLU}$), and $[ \cdot \| \cdot ]$ denotes the concatenation operator.  A well-trained Q-function should assign high values to feasible assembly actions while effectively filtering out infeasible ones.

\subsection{Q-Function Training} \label{sec:q_function_training}
For each state-action pair $(s_t, a_t)$ in the MDP, consider a reward function $R(s_t, a_t)$ that assigns a positive reward ($+1$) when $a_t$ successfully transitions $s_t$ to the full assembly state $v_N$, a negative reward ($-1$) if $a_t$ corresponds to a physically infeasible assembly action from $s_t$, and a neutral reward ($0$) otherwise. However, as is common with the sparse reward problem widely discussed in the Q-learning literature, this reward function design may unintentionally guide the robot to select an action that leads to a dead end -- a node with no feasible edges to its out-neighbors in $\mathcal G$.

To address this issue, we propose a delayed reward assignment strategy. During the training process, when the state $s_t$ transitions to the next state $s_{t+1}$ under an assembly action $a_t$, the RL agent delays assigning a reward until it confirms whether a feasible sequence of actions exists from $s_{t+1}$ to the full assembly state $v_N$. If no such sequence is found, the episode terminates, and a reward of $-1$ is given. Otherwise, a reward of $0$ is assigned. This approach ensures that when selecting an action, both its physical feasibility (verified through the physics-based simulation) and its impact on the remaining assembly process are considered. Additionally, the proposed strategy improves sample efficiency by preventing excessive accumulation of experiences leading to dead ends in the replay buffer, which could otherwise lower the occurrence of positive samples.

To implement the delayed reward assignment strategy, it is necessary to perform feasibility checks for the remaining assembly actions beyond $s_{t+1}$. This could lead to redundant evaluation of the same assembly actions within each epoch for a single assembly problem instance. To avoid such inefficiency, in our implementation, we continuously update the label $y_{vu}$ for each evaluated action $(v,u)$ in $\mathcal G$, eliminating the need for re-evaluating the feasibility of assembly actions. Algorithm~\ref{alg:cap} outlines the implementation of this strategy.
It is important to note that the information required for the delayed reward assignment strategy -- specifically, checking the existence of feasible actions leading to $v_N$ -- is only necessary during the training phase. During testing, such checks are no longer required.

With all the components defined for training the GNN, we employ a Double Deep Q-Network (DDQN) \cite{van_Hasselt_Guez_Silver_2016} to learn a Q-function that predicts the likelihood that a selected assembly action is physically feasible and leads to the full assembly state $v_N$. The Q-function's parameters $\theta = \{\mathbf{a}^{(1)}, \dots, \mathbf{a}^{(L)}, \mathbf{\Theta}_1^{(1)}, \dots, \mathbf{\Theta}_4^{(L)} \}$ are updated by minimizing the following loss function: $L = \left(y_t - Q\left(s_t, a_t ; \theta \right)\right)^2$, where $y_t = r_t + \gamma Q^{\prime}\left(s_{t+1}, \operatorname{argmax}_a Q\left(s_{t+1}, a; \theta\right); \theta^- \right)$ uses a target network $Q^{\prime}$ to evaluate the action selected by our evaluation function $Q$. The target network's parameters $\theta^-$ are periodically updated by copying $\theta$ from $Q$.

\begin{algorithm}
\caption{Delayed reward assignment strategy}\label{alg:cap}
    \DontPrintSemicolon
    \SetKwFunction{FCheckFeasibility}{CheckFeasibility}
    \SetKwProg{Fn}{Function}{:}{}
    \SetKwFunction{proc}{RewardAssignment}
    \SetKwProg{myproc}{Procedure}{:}{}

    \Fn{\FCheckFeasibility{$(v, u)$}} {
        Check if a feasible edge exists from $v$ to $u$ using RRT$^*$ and Pybullet, and return the result ($0$/$1$).
    }

    \myproc{\proc{$(v,u), \mathcal G$}} {
    \If{$y_{vu} = none$} {
        \tcp*[l]{\footnotesize{if $y_{vu}$ has not yet been assessed}}
        $y_{vu} \gets $ \FCheckFeasibility{$(v, u)$}
    }
    
    \If{$y_{vu} = 0$ } {
        \tcp*[l]{\footnotesize{infeasible action}}
    
        reward $\gets -1$, done $\gets true$ 
    }
    \uElseIf{$y_{vu} = 1$ and $u = v_N$} {
        \tcp*[l]{\footnotesize{complete sequence}}
        
        reward $\gets +1$, done $\gets true$ 
    }
    \Else {
        reward $\gets -1$
    
        $v' \gets u$
        
        \For{$u' \in \mathcal{N}_{+}(v')$} {
            future\_reward, done $\gets$ \proc{$(v,u), \mathcal G$}

            \If{future\_reward $=0$ or future\_reward $=1$} {
                reward $\gets 0$
                
                break
            }
        }
    }   
    
    \If {reward $= 1$ or reward $= -1$} {
        done $\gets true$
    }
    \Else {
        done $\gets false$
    }
    
    \Return reward, done    
}
\end{algorithm}

\subsection{Action Selection}
After training the GNN, we use the learned Q-function to identify valid paths in the directed graph representing an assembly problem instance. Note that during the testing phase, the feasibility of each assembly action is not directly available. The S2A framework applies a greedy policy to search for the optimal sequence of assembly actions by selecting the edge with the highest Q-value at each step. The feasibility of the chosen action is then verified using the PyBullet simulation and RRT$^*$. If the action is physically feasible, the MDP state transitions to the corresponding next state. This process continues until the full assembly state $v_N$ is reached or an infeasible assembly action is selected. 

Similar to other graph-based approaches for CO problems \cite{NIPS2015_29921001, NEURIPS2018_9fb4651c, joshi2019efficient}, the S2A can be enhanced with advanced inference techniques, such as beam search, to improve solution quality by generating multiple candidate solutions for the ASP problem. Starting from the root $v_1$, we iteratively expand the $b$ most feasible assembly sequences at each step until reaching the full assembly state $v_N$. Here, $b$ is the beam width. The resulting top-$b$ candidates are verified one by one in the simulation.
We demonstrate the effectiveness of this approach through experiments in Section~\ref{sec:experiments}.

\section{Experiments} \label{sec:experiments}
Through experiments, we aim to: 1) evaluate the effectiveness of our method compared to baseline approaches across assembly problems of four different sizes, 2) assess the generalizability of our method to unseen assembly problem instances, and 3) demonstrate its performance in real-world experiments using a robotic manipulator.

\subsection{Experimental Setup}
\subsubsection{Environments}
To define the simulation environments, we generate $K$ assembly problem instances, where each part's geometry is derived from unit cubes, with all these instances assembled into a regular $3 \times 3 \times 3$ structure. In constructing each part, the positions of the unit cubes are randomly selected, while ensuring that each cube is in contact with at least one other cube, resulting in a well-defined geometry. This process is repeated for four different problem sizes, $M \in \{4,5,6,7\}$, with the following number of instances per size: $K = 272, 301, 483, 728$. Throughout this process, we exclude simple structures, particularly parts containing only $1$ or $2$ unit cubes, by ensuring each part has a comparable number of cubes. 
During training and testing, an assembly problem instance, defined by its associated graph $\mathcal{G}$ as described in Section~\ref{sec:formulation}, is sampled at each episode. Using DDQN, assembly sequence data are collected, and the network parameters are updated following the delayed reward assignment strategy outlined in Algorithm~\ref{alg:cap}.

\subsubsection{Implementation Details}
We utilize Open Motion Planning Library \cite{ompl} for the RRT* planner (with a search budget of $400$ seconds), PyTorch Geometric \cite{pyg} to implement two GATv2 layers ($L=2$) with eight attention heads (hidden size of $128$), and Tianshou \cite{tianshou} for training and testing the DDQN algorithm using the Adam optimizer (batch size of $32$, learning rate of $10^{-4}$). The discount factor $\gamma$ is set to $0.99$, and the target network's parameters are updated after every $100$ training steps. We train the GNN model with and without the delayed reward assignment strategy for $2,000$ epochs, using $5$ distinct seeds. 

\subsubsection{Baseline Methods}
During testing, we employ beam search with width $b \in \{1,3,5\}$ to construct solutions to randomly sampled problem instances for every problem size $M$. Note that beam search with $b=1$ is equivalent to the standard greedy search. For performance comparison, we select the following three types of baselines: 1) \textit{RandomWalk} is a stochastic policy that starts at the root $v_1$ and moves to one of its out-neighbors with equal probability at each step. 2) \textit{Heuristic Algorithm} which prioritizes assembling parts with a lower center of mass, based on the commonly observed tendency for humans to assemble bottom parts first. 3) A GNN-based method without the delayed reward assignment strategy, which utilizes the reward function described at the beginning of Section~\ref{sec:q_function_training}.\footnote{Various GNN-based approaches have been explored in the robotic assembly literature. However, since one of our key contributions is the new strategy for GNN training, rather than the GNN architecture itself, we ensure a fair comparison by evaluating the strategy using the same GNN architecture as in our S2A framework.} All methods are tested in random environments that cover the entire cube dataset.

\subsection{Results}

\begin{table}
\centering
% \scriptsize
\caption{Comparison of the proposed method's performance against baseline approaches (BS1/3/5: beam search with $b=1,3,5$).}
\begin{tabularx}{\linewidth}{cc|>{\centering\arraybackslash}X>{\centering\arraybackslash}X>{\centering\arraybackslash}X>{\centering\arraybackslash}X}
\toprule
\multicolumn{2}{c|}{\multirow{2}{*}{\textbf{Method}}} & \multicolumn{4}{c}{\textbf{Success Rate (\%)}} \\
                                      &                     & M=4   & M=5   & M=6   & M=7  \\ 
\midrule
\multirow{5}{*}{\textbf{Baseline}}    & RandomWalk          & 8.15  & 3.81  & 1.33  & 0.62  \\
                                      & Heuristics          & 45.60 & 34.87 & 21.02 & 16.67 \\
                                      & GNN (w/o delay), BS1 & 87.72 & 78.91 & 59.50 & 59.72 \\
                                      & GNN (w/o delay), BS3 & 96.17 & 89.04 & 74.04 & 74.39 \\
                                      & GNN (w/o delay), BS5 & 96.90 & 91.38 & 77.78 & 76.97 \\
\midrule
\multirow{3}{*}{\textbf{Ours}}        & S2A, BS1            & 91.50 & 87.85 & 70.00 & 72.43 \\ 
                                      & S2A, BS3            & 98.29 & 93.90 & 81.29 & 82.15 \\ 
                                      & \textbf{S2A, BS5}   & \textbf{99.09} & \textbf{95.13} & \textbf{84.01} & \textbf{84.71} \\
\bottomrule
\end{tabularx}
\label{tab:1}
\vspace{0em}
\end{table}

\begin{figure}
    \centering
    \subfigure[\label{fig:beam}]{\includegraphics[width=0.49\linewidth]{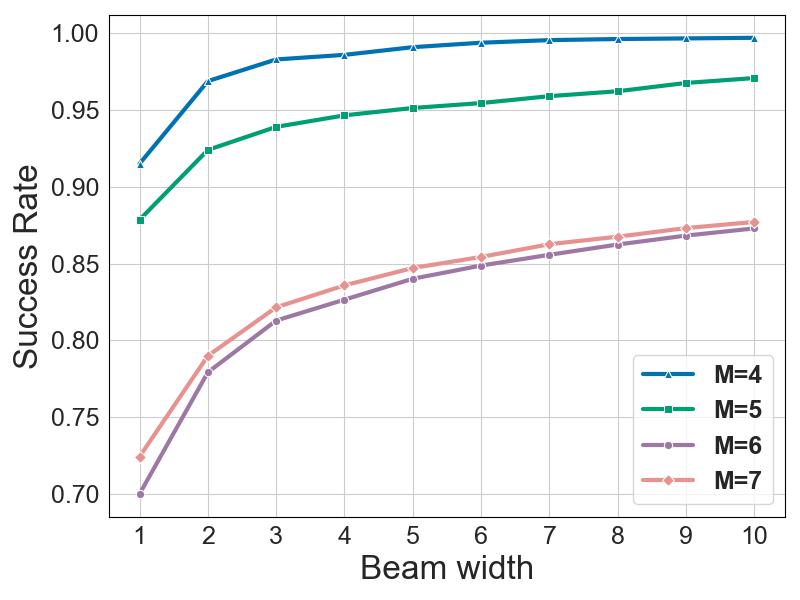}}
    \subfigure[\label{fig:curves}]{\includegraphics[width=0.49\linewidth]{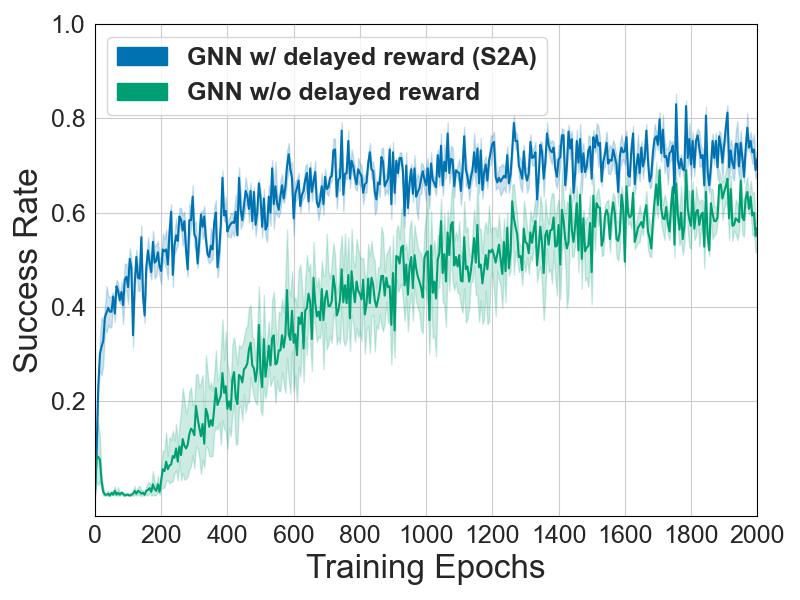}}
    % \subfigure[\label{fig:beam}]{\includesvg[width=0.49\linewidth]{figs/beam.svg}}
    % \subfigure[\label{fig:curves}]{\includesvg[width=0.49\linewidth]{figs/cube_7_parts.svg}}
    \caption{(a) A graph illustrating the success rate improvement of our method as the beam width increases, and (b) Comparison of training curves between two GNN-based approaches for the $7$-part assembly problem.}
    \vspace{-1em}
\end{figure}

We evaluate the proposed method against the baseline methods based on the \textit{success rate}, defined as the percentage of test cases where the S2A successfully finds an assembly sequence from $v_1$ to the full assembly state $v_N$. Table~\ref{tab:1} summarizes the simulation results. As shown, our S2A method significantly outperforms the baseline methods. Both non-learning-based methods suffer greatly from the combinatorial explosion issue. In contrast, the S2A maintains a high success rate of over 70\% across all assembly problems, even when using beam search with $b=1$. Interestingly, the performance of our method improves slightly as the problem size increases from 6 to 7. We speculate that this is due to the simpler geometries of the individual parts in larger problems, as each part consists of fewer unit cubes, making it easier for our method to learn underlying precedence relationships. Additionally, using beam search improves the success rate by generating multiple solution candidates, with performance gains increasing as the beam width $b$ grows, as illustrated in Fig.~\ref{fig:beam}. However, since the computation time for validating multiple candidates increases with $b$, it is important to find an optimal trade-off between $b$ and computation time.

Fig.~\ref{fig:curves} compares the training results of two GNN-based methods -- both using the same GNN architecture, with and without the delayed reward assignment strategy. As observed, the proposed strategy leads to faster convergence and higher performance.\footnote{Although the delayed reward assignment strategy generally takes more time to assign rewards to the RL agent, resulting in longer total training time, it requires less number of training updates. This is because parameter updates occur only when a fixed amount of new data becomes available, allowing for more efficient learning during training.} In contrast, the GNN method without the strategy requires a large number of exploration steps before showing performance improvements and tends to converge at a lower performance level. These observations highlight the effectiveness of the delayed reward assignment strategy.

We also evaluate the generalization of our approach to assembly problem instances of varying sizes, demonstrating its ability to solve problems with different numbers of parts using a single GNN model. To assess this generalizability, we train the S2A model on problem instances of one size and test it on instances of other sizes, following the one-to-many validation method used in \cite{10342352}. Based on Fig.~\ref{fig:beam}, we use a fixed beam width of $b=3$, as increasing the width beyond this point produces only a marginal benefit. As shown in Table~\ref{tab:generalization}, our method generalizes well to smaller problem instances, where the combinatorial complexity is lower than that of the training instances. However, performance degradation is observed when applied to larger problem instances. Despite this, our method still achieves reasonable performance, consistently outperforming the two non-learning-based baselines.

\begin{table}[htbp]
\centering
% \scriptsize
\caption{Results of the generalization test using beam search ($b=3$).}
\begin{tabularx}{0.8\linewidth}{>{\centering\arraybackslash}X|>{\centering\arraybackslash}X>{\centering\arraybackslash}X>{\centering\arraybackslash}X>{\centering\arraybackslash}X}
\toprule
Test $\rightarrow$ & \multicolumn{4}{c}{\textbf{Success Rate (\%)}} \\
Train $\downarrow$ &  M=4  &  M=5  &  M=6  & M=7  \\ 
\midrule
M=4                &   -   & 80.14 & 61.98 & 54.95 \\
M=5                & 91.38 &   -   & 67.61 & 63.01 \\
M=6                & 92.41 & 84.36 &   -   & 72.88 \\ 
M=7                & 92.93 & 86.21 & 73.47 &   -   \\
\bottomrule
\end{tabularx}
\label{tab:generalization}
\vspace{-1em}
\end{table}

\subsection{Robot Demonstration}
\begin{figure}
    \centering
    \includegraphics[width=\linewidth, height=11em]{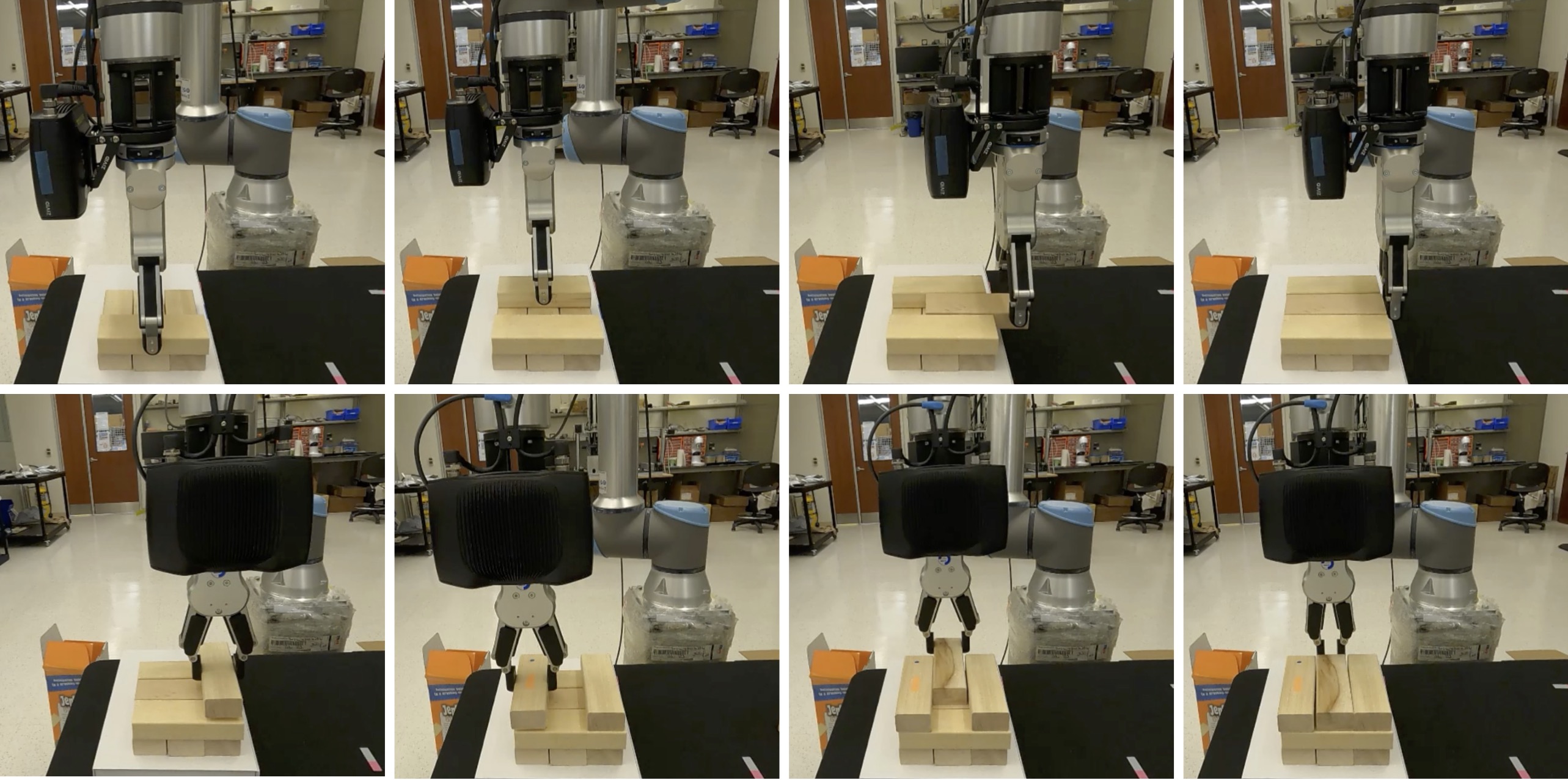}
    \caption{Real-world execution of an assembly sequence planned by the proposed S2A framework. Assembling the middle part involves both grasping and pushing.}
    \label{fig:jenga-demo}
    \vspace{-1em}
\end{figure}

Fig.~\ref{fig:jenga-demo} depicts a demonstration of Jenga assembly using a robotic manipulator. Given a set of Jenga blocks along with their designated assembly locations, we derive the volumetric representations for all subassemblies of the Jenga tower. At every assembly step, we pass the representations as input to the S2A framework to compute the Q-values for all admissible assembly actions. We apply beam search ($b=1$) to find a feasible assembly sequence.

\section{Conclusion} \label{sec:conclusion}
We introduced the S2A framework for efficient ASP, leveraging graph-based RL. Extensive simulations confirmed the effectiveness of our approach, which outperformed all baseline methods and demonstrated strong generalization to unseen problem instances of various sizes. We also demonstrated its performance in a real-world experiment.
As a future direction, our method can be extended to real-world object assembly by incorporating CAD models. Additionally, to address the current limitation of adding only a single part to an existing subassembly, future enhancements could focus on enabling parallel execution of large-scale assemblies through a divide-and-conquer strategy.

% \addtolength{\textheight}{-12cm}   % This command serves to balance the column lengths
                                  % on the last page of the document manually. It shortens
                                  % the textheight of the last page by a suitable amount.
                                  % This command does not take effect until the next page
                                  % so it should come on the page before the last. Make
                                  % sure that you do not shorten the textheight too much.

%%%%%%%%%%%%%%%%%%%%%%%%%%%%%%%%%%%%%%%%%%%%%%%%%%%%%%%%%%%%%%%%%%%%%%%%%%%%%%%%

%%%%%%%%%%%%%%%%%%%%%%%%%%%%%%%%%%%%%%%%%%%%%%%%%%%%%%%%%%%%%%%%%%%%%%%%%%%%%%%%

%%%%%%%%%%%%%%%%%%%%%%%%%%%%%%%%%%%%%%%%%%%%%%%%%%%%%%%%%%%%%%%%%%%%%%%%%%%%%%%%

% Appendixes should appear before the acknowledgment.

% \section*{ACKNOWLEDGMENT}

%%%%%%%%%%%%%%%%%%%%%%%%%%%%%%%%%%%%%%%%%%%%%%%%%%%%%%%%%%%%%%%%%%%%%%%%%%%%%%%%
\balance
\bibliography{ref}

% \section*{APPENDIX}
\newpage
\appendix
\subsection{Object Manipulation Actions and Trajectory Planning} \label{sec:object_manipulation}
\begin{figure} [t!]
    \centering
    \subfigure[\label{fig:object_manipulation_a}]{\includegraphics[width=0.32\linewidth]{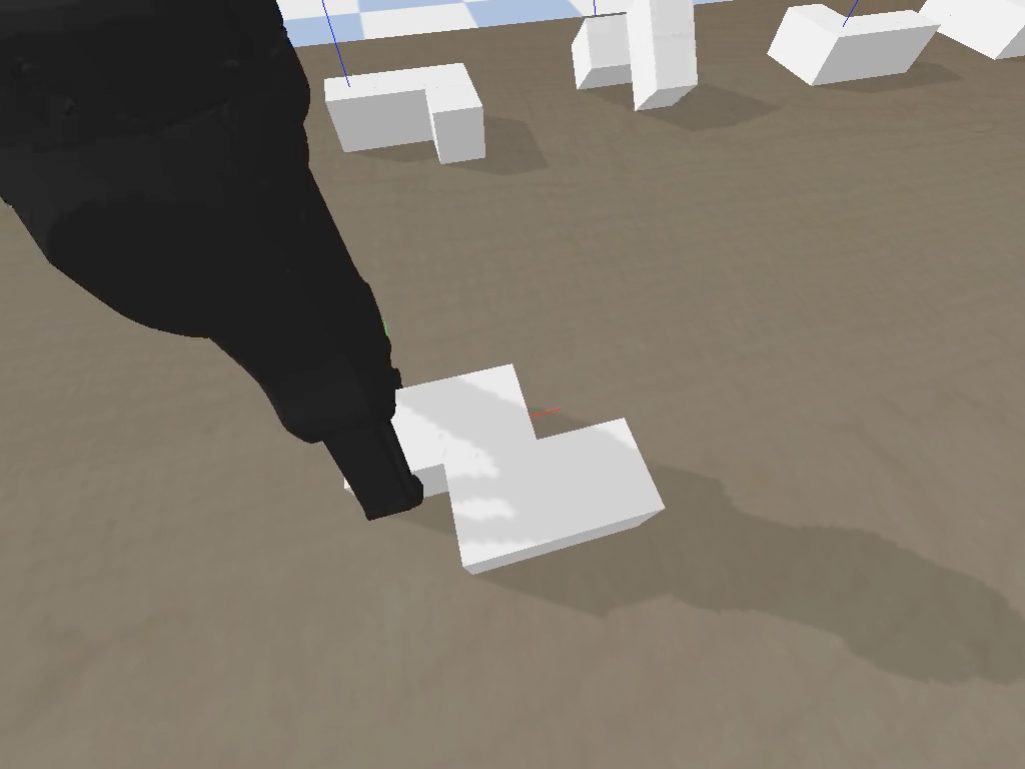}}
    \subfigure[\label{fig:object_manipulation_b}]{\includegraphics[width=0.32\linewidth]{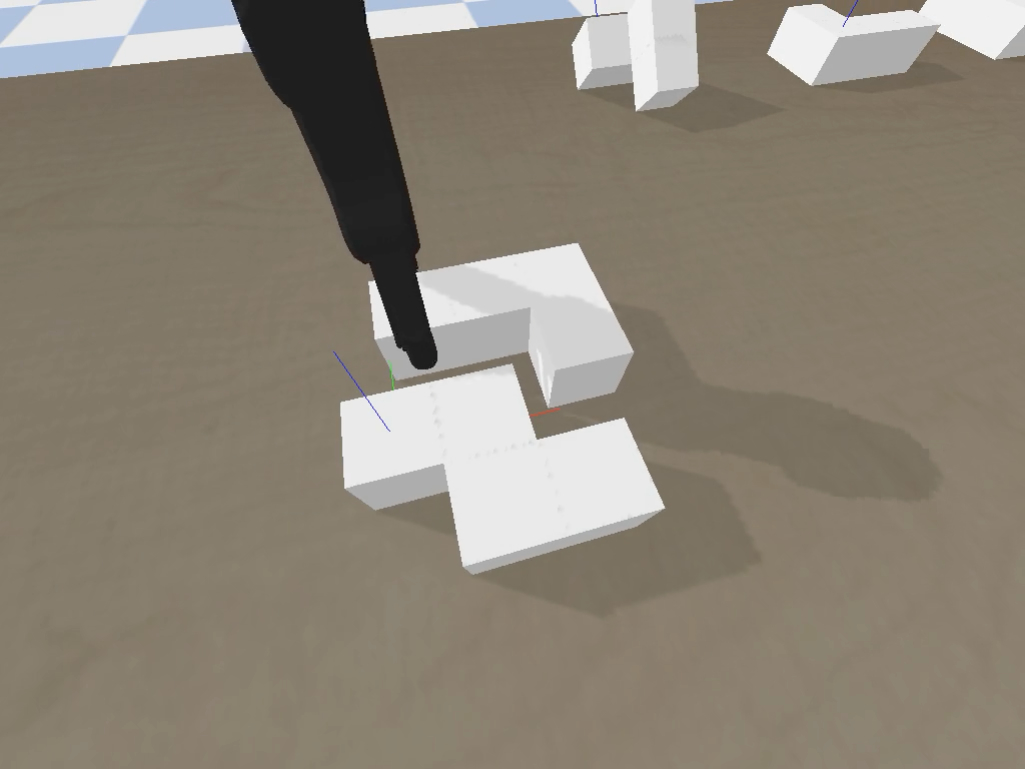}}
    \subfigure[\label{fig:object_manipulation_c}]{\includegraphics[width=0.32\linewidth]{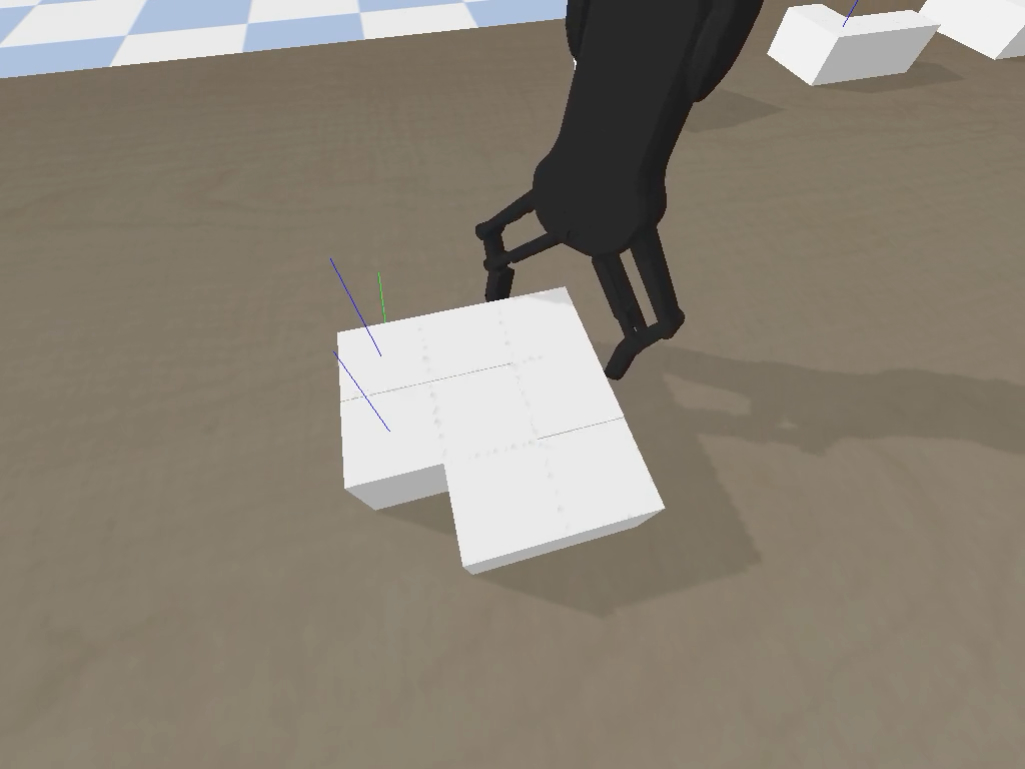}}
    \caption{An illustration of object manipulation actions: grasping (a),(b) and pushing (c).}
    \label{fig:object_manipulation}
\end{figure}

We employ two object manipulation actions for the assembly process: grasping and pushing, as illustrated in Fig.~\ref{fig:object_manipulation}. For grasping, the locations are predefined at the centers of two opposite sides of each unit cube constituting the part, with grasping orientations sampled in multiples of $90^\circ$. Using PyBullet simulation, we check for potential collisions between the gripper and each part, and only select collision-free grasping locations and orientations. 

\begin{figure} [t!]
    \centering
    \includegraphics[width=0.6\linewidth]{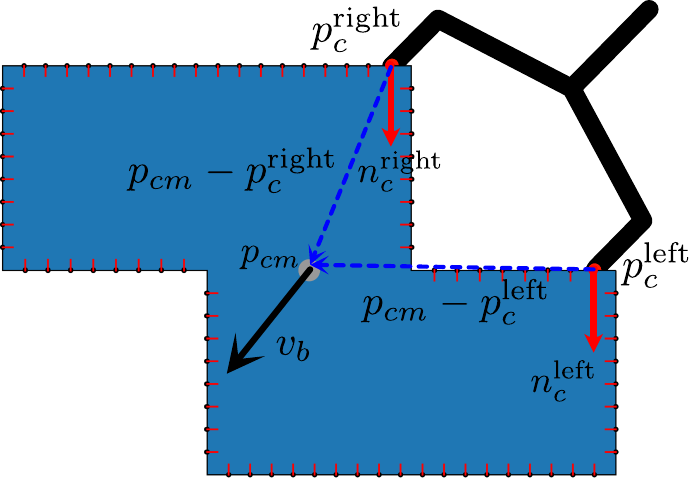}
    \caption{An illustration of the process for selecting two contact points for object pushing.}
    \label{fig:push_contact_selection}
\end{figure}

For pushing, the action is constrained to occur within a 2D plane parallel to the floor. To execute this, we sample the horizontal plane containing the part's center of mass $p_{cm}$ and search for two contact points $p_c^{\text{left}}$ and $p_c^{\text{right}}$, illustrated in Fig.~\ref{fig:push_contact_selection}, satisfying the following three criteria:
\begin{enumerate}
\item The distance between $p_c^{\text{left}}$ and $p_c^{\text{right}}$ is less than the maximum gripper opening. There is no collision between the gripper and part when the tips of the gripper touch the contact points.

\item The inward normal directions $n_c^{\text{left}}$ and $n_c^{\text{right}}$ are strictly positively correlated with the desired moving direction $v_b$ of the part. Specifically, it requires
\begin{align*}
    n_c^{\text{left}} \cdot v_b > 0 \quad \text{and} \quad n_c^{\text{right}} \cdot v_b > 0.
\end{align*}
This ensures that slippage between the gripper and the contacting surfaces of the part is minimized.

\item The forces applied at the two contact points generate rotational motion of the part. To prevent the part from rotating during the push, we impose the following condition to ensure opposite rotational directions at the two contact points:
\begin{align*}
    \left( (p_{cm} - p_c^{\text{left}}) \times v_b \right) \left( (p_{cm} - p_c^{\text{right}}) \times v_b \right) < 0,
\end{align*}
where $p_{cm}$ is the center of mass, and $v_b$ is the desired moving direction.
\end{enumerate}
Once the two contact points are selected, the gripper moves along the desired velocity $v_b$ to push the part toward its intended location.

Using these two object manipulation actions, we proceed in PyBullet simulation to find collision-free motion trajectories for both the part and the gripper:
\begin{enumerate}
    \item First, we verify if the gripper can grasp the part at its designated assembly location using one of the pre-defined grasping configurations (Fig.~\ref{fig:object_manipulation_a}).
    If grasping at the assembly location is feasible, we use the RRT$^\ast$ planner to determine whether a collision-free trajectory exists for the selected grasping configuration to reach the assembly location from an initial location of the part. We iterate through all feasible grasping configurations until a collision-free trajectory is found.

    \item If no feasible grasping configuration with a collision-free trajectory is found, considering the dimensions of the gripper tips, we slightly adjust the assembly location without altering its orientation (Fig.~\ref{fig:object_manipulation_b}). If a new location where the gripper can grasp the part is identified, we then check if two valid contact points can be established and the gripper can push the part to the original assembly location without colliding with other parts (Fig.~\ref{fig:object_manipulation_c}).
    
\end{enumerate} 

When planning the trajectory, priority is given to grasping-only manipulation, as it tends to be more reliable than a combination of grasping and pushing for assembly tasks.

\end{document}